\documentclass[twocolumn]{article}
\usepackage{epsfig}
\newtheorem{mystrategy}{Strategy}
\newenvironment{strategy}[1][]{\begin{mystrategy}[#1]\sl}{\end{mystrategy}}

\newtheorem{mytheorem}{Theorem}

\begin{document}
\bibliographystyle{plain}
\thispagestyle{empty}

\title{A Fundamental Algorithm for Dependency Parsing \\ \textsf{\textbf{(with corrections)}}}

\author{
Michael A. Covington\\
\quad\\
        \textsf{\textbf{Institute for Artificial Intelligence}}\\
        The University of Georgia\\
        Athens, GA 30602-7415 U.S.A.\\
        mc@uga.edu
}

\date{\quad}

\maketitle

\insert\footins{\vspace*{0.9in}}    

{\small
\parskip=12pt
\parindent=0pt

{\bf Abstract--}
This paper presents a fundamental algorithm for parsing natural
language sentences into dependency trees.
Unlike phrase-structure (constituency) parsers, this algorithm operates one word at
a time, attaching each word as soon as it can be attached, corresponding to properties
claimed for the parser in the human brain.
Like phrase-structure parsing, its worst-case complexity is $O(n^3)$, but in human
language, the worst case occurs only for small $n$.
}

\parindent=2em

\section{Overview.}

This paper develops, from first principles,
several variations on a fundamental algorithm for parsing natural
language into dependency trees.
This is an exposition of an algorithm that has been known, in some form,
since the 1960s but is not presented systematically in the extant literature.%
\footnote{This paper first appeared in
\emph{Proceedings of the 39th Annual ACM Southeast Conference}
(2001), ed. John A. Miller and Jeffrey W. Smith, pp. 95--102.
Copyright 2001 Association for Computing Machinery (www.acm.org).
Further publication requires permission.
\textsf{\textbf{This version of the paper, released August 2010, contains corrections,
which are set in bold sans-serif type.  I thank W.\ Cody Boisclair for pointing
out an error in the original version of one of the algorithms.}}
}

Unlike phrase-structure (constituency) parsers,
this algorithm operates one word at a time, attaching each word as soon as it can be
attached.
There is good evidence that the parsing process used by the human mind has these
properties \cite{Abney}.

\section{Dependency grammar.}

\subsection{The key concept.}

\begin{figure}
\centerline{\includegraphics[width=3in]{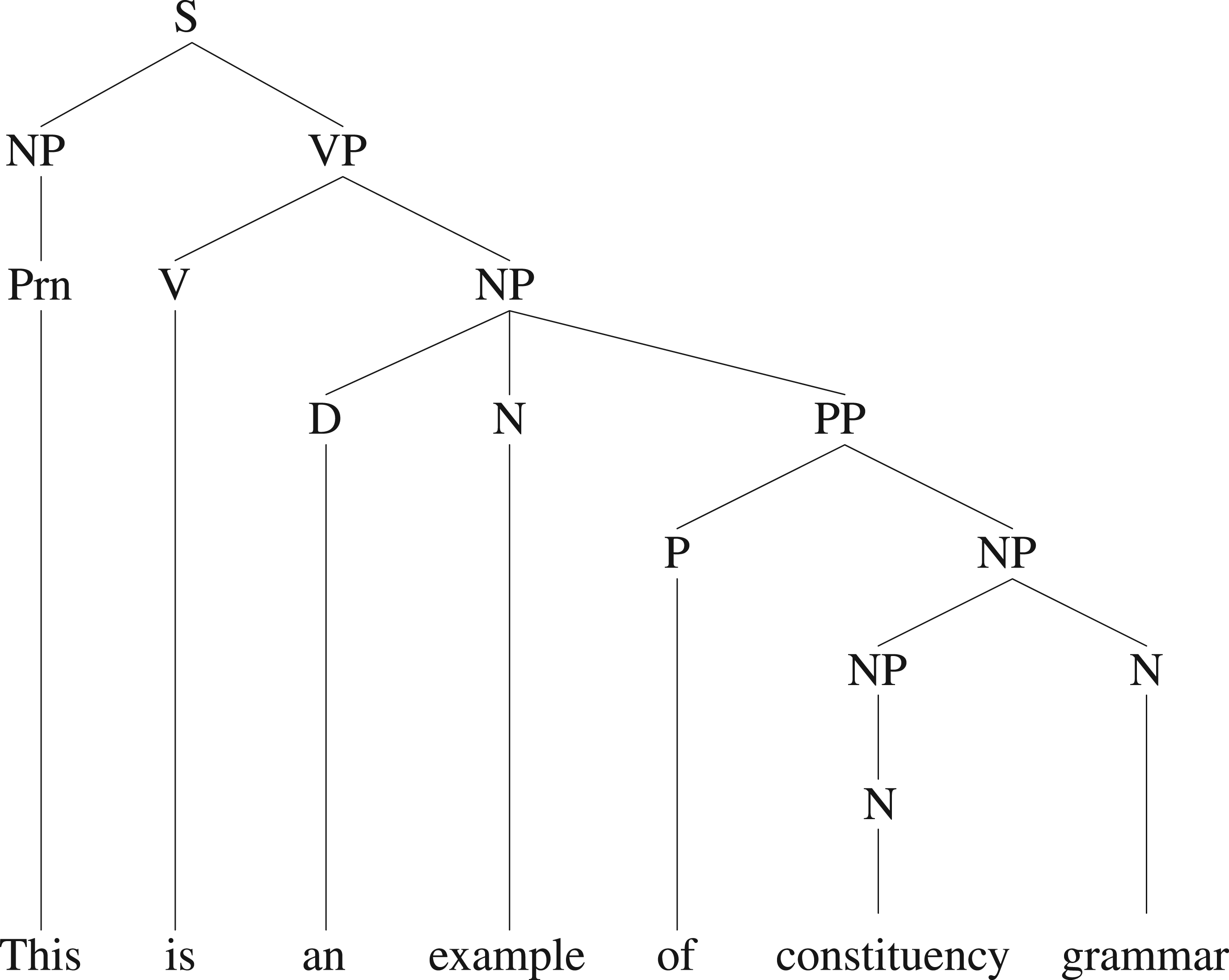}}
\caption{A constituency tree.}
\label{consex}
\end{figure}

\begin{figure}
\centerline{\includegraphics[width=3in]{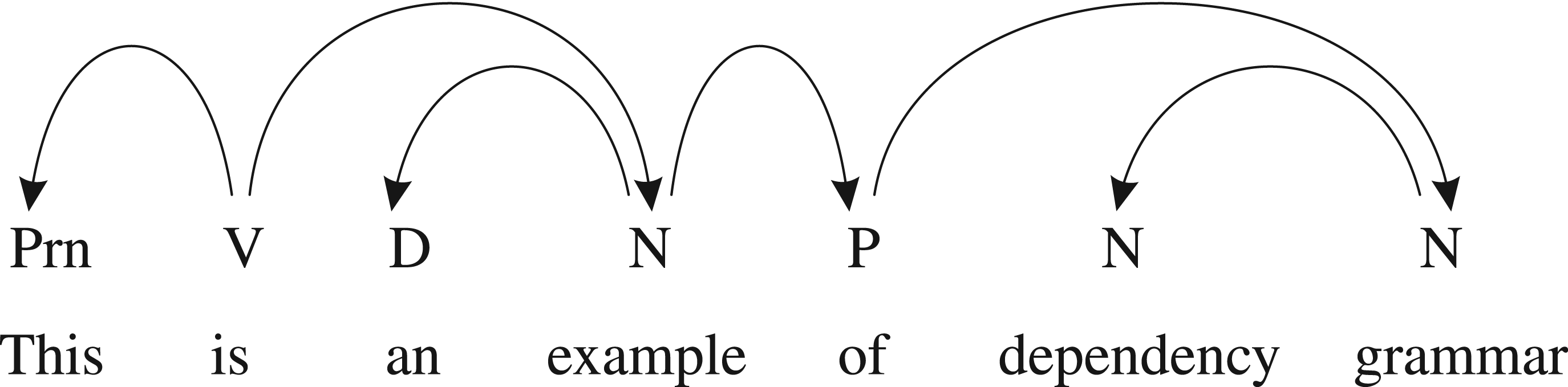}}
\vspace{0.25in}
\centerline{\includegraphics[width=1.3in]{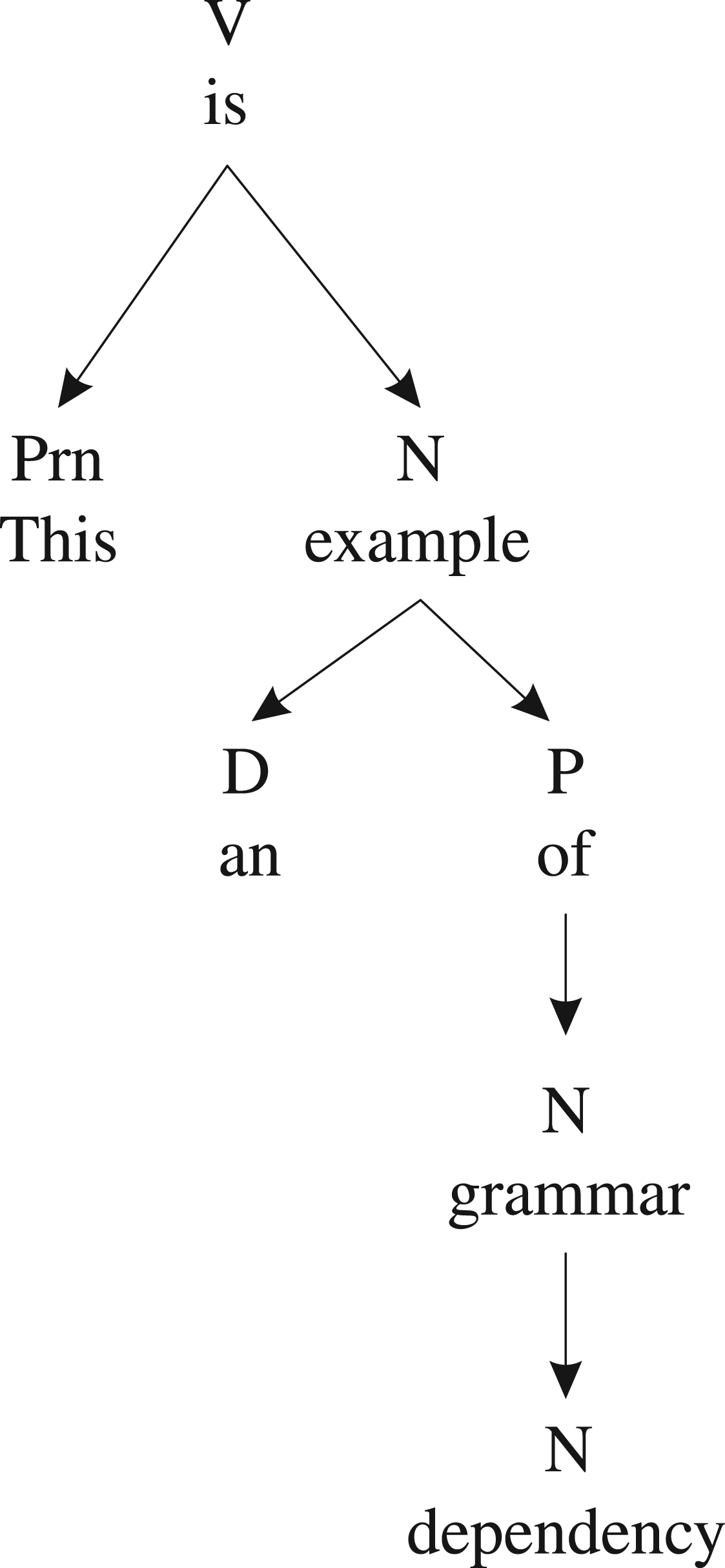}}
\vspace{0.25in}
\caption{A dependency tree is a set of links connecting heads to dependents.}
\label{deptree}
\end{figure}

There are two ways to describe sentence structure in natural language: by
breaking up the sentence into \textbf{constituents} (phrases), which are
then broken into smaller constituents (Fig.~\ref{consex}), or by drawing links
connecting individual words (Figs.~\ref{deptree}, \ref{depex}).  These are called
\textbf{constituency grammar} and \textbf{dependency grammar} respectively.

Constituency grammar appears to have been invented only once,
by the ancient Stoics \cite{Mates61}, from whom it was passed through formal logic
to linguists such as Leonard Bloomfield, Rulon Wells, Zellig Harris, and Noam Chomsky.
It is also the basis of formal language theory as studied by computer scientists.

Dependency grammar, on the other hand, has apparently been invented many
times and in many places.  The concept of a word-to-word link occurs naturally
to any grammarian who wants to explain agreement, case assignment, or any
semantic relation between words.  Dependency concepts are found in traditional Latin,
Arabic, and Sanskrit grammar, among others.  Computer implementations of dependency
grammar have attracted interest for at least 40 years \cite{HayZie, Hays64, Hays66, Fras93,
Fras94, Sgall},
but there has been little
systematic study of dependency parsing, appparently due to the widespread misconception
that all dependency parsers are notational variants of constituency parsers.

\subsection{Dependency trees.}

Whenever two words are connected by a dependency relation, we say that
one of them is the \textbf{head} and the other is the \textbf{dependent},
and that there is a \textbf{link} connecting them.
In general, the dependent is the modifier, object, or complement; the head
plays the larger role in determining the behavior of the pair.
The dependent presupposes the presence of the head; the head may require
the presence of the dependent.

Figure \ref{deptree} shows the dependency structure
of a sentence.
Essentially, a dependency link is an arrow pointing from head to dependent.
The dependency structure is a tree (directed acyclic graph) with the main
verb as its root (head).

\begin{figure}
\centerline{\includegraphics[width=3in]{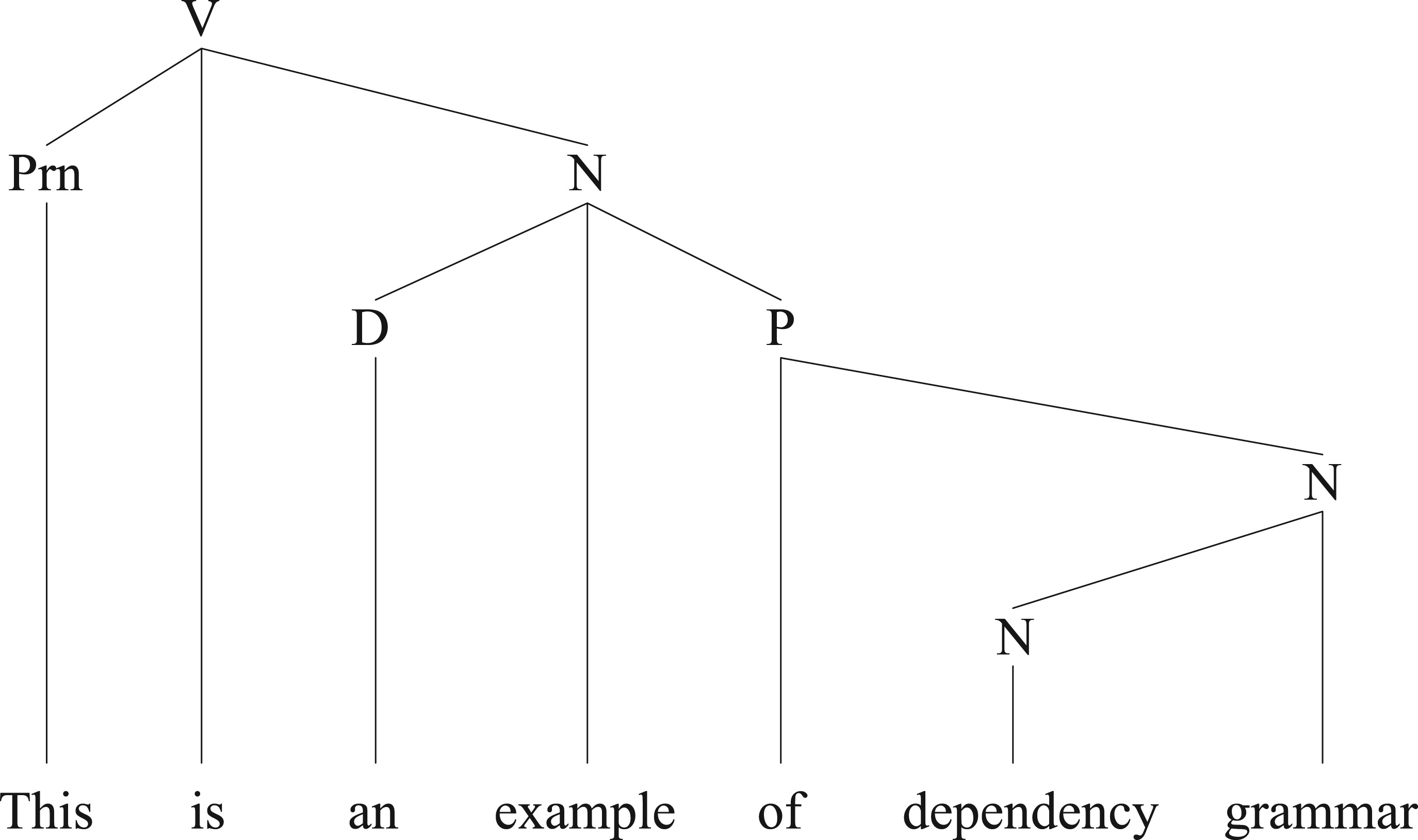}}
\caption{This representation of a dependency tree preserves the word order
while depicting the tree structure plainly.  To get from a head to its
dependents, go downhill.}
\label{depex}
\end{figure}

Figure \ref{depex} shows a way to display the word order and the tree structure
at once.  To get from a word to its dependents in this kind of diagram, go downhill.

In what follows, a dependent that precedes its head is called a \textbf{predependent}; one that
follows its head, a \textbf{postdependent}.

I shall say that a word is \textbf{independent} (headless) if it is not a dependent
of any other word.

Note that in the dependency tree, constituents (phrases) still exist.  Any word
and all its dependents, their dependents, etc., form a phrase.  I shall say
that dependents, dependents of dependents, etc., are \textbf{subordinate} to
the original word, which in turn \textbf{dominates} (is \textbf{superior} to) them.

A word \textbf{comprises} itself and all the words that it dominates.
That is, the head of a phrase comprises the whole phrase.

\subsection{Generative power.}

In 1965, Gaifman \cite{Gaif65} proved that dependency grammar and constituency
grammar are \textbf{strongly equivalent} --- that they can generate the same sentences
and make the same structural claims about them --- provided the constituency grammar
is restricted in a particular way.  The restriction is that one word in each phrase
is designated its \textbf{head}, and the phrase has no name or designation apart
from the designation of its head.

\begin{figure}
\vspace*{0.25in}
\centerline{\includegraphics[width=3in]{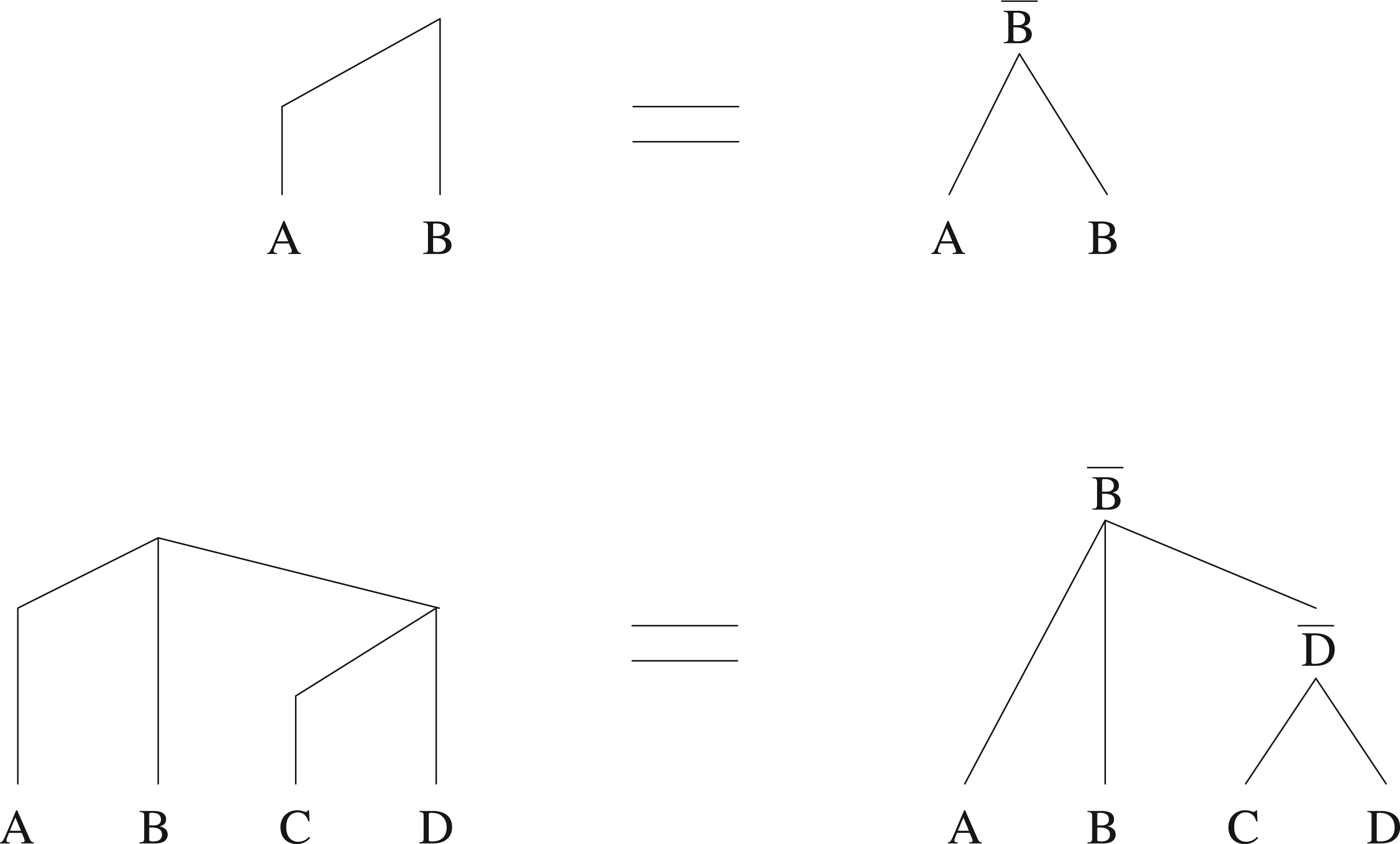}}
\caption{Equivalent dependency and constituency trees.}
\label{equiv}
\end{figure}

That is tantamount to saying that a noun phrase
has to be built around a noun, a verb phrase around a verb, and so forth.
Just take the category name ``noun'' or ``verb,'' add ``phrase,'' and you have the
name of the phrase that it heads.
(In a pure constituency grammar, NP and VP are atomic symbols not related to N and V,
a fact all too seldom appreciated.)

Linguists
have accepted this proposed restriction for
other reasons; they call it \textbf{X-bar theory}
\cite{Jack77}.  Thus, constituency grammar as currently practiced is very close to being
a notational variant of dependency grammar.  Figure \ref{equiv} shows interconversion
of dependency and constituency trees.  A bar over a category label indicates that it
labels a phrase rather than a word.

\subsection{The appeal of dependency parsing.}

In what follows I shall explore some parsing algorithms that use the dependency
representation.  Please note that I am not claiming any significant difference
in generative power between dependency grammar and constituency grammar; still less
am I claiming that English, or any other human language, ``is a dependency language''
rather than a constituency language, whatever that might mean.
Nor do I address any technical aspects of
constructing an adequate dependency grammar of English.  My concern is only the
formalism.
\emph{Prima facie,} dependency parsing offers some advantages:
\begin{itemize}
\item Dependency links are close to the semantic relationships needed for the next
stage of interpretation; it is not necessary to ``read off'' head-modifier or
head-complement relations from a tree that does not show them directly.
\item The dependency tree contains one node per word.  Because the parser's job
is only to connect existing nodes, not to postulate new ones, the task of parsing
is in some sense more straightforward.  (We will presently see that the actual order
of complexity is no lower, but the task is nonetheless easier to manage.)
\item Dependency parsing lends itself to word-at-a-time operation, i.e., parsing
by accepting and attaching words
one at a time rather than by waiting for complete phrases.

Abney \cite{Abney} cites several kinds of evidence that the
parser in the human mind operates this way.  Consider for example a verb phrase
that may or may not contain a direct object,
such as \emph{sang loudly} (vs.~\emph{sang songs loudly}).
A top-down constituency
parser has to choose \emph{a priori} whether to expect the object or not,
before it has any way to know which choice is right, and then has to
backtrack if it guessed wrong;
that is \emph{spurious local ambiguity,} apparently absent in human parsing.
A bottom-up constituency parser cannot construct the verb phrase
until all the words in it have been encountered; yet people clearly begin to understand
verb phrases before they are over.
My dependency parser has neither problem; it accepts words and attaches them with
correct grammatical relations as soon as they are encountered, without making any
presumptions in advance.

\end{itemize}


\section{The parsing task.}

The task of a dependency parser is to take a string of words and impose on it
the appropriate set of dependency links.  In what follows I shall make several
assumptions about how this is to be done.

\subsection{Basic assumptions.}

\begin{itemize}
\item
\textbf{Unity}: The end product of the parsing process is a single tree (with a
unique root) comprising all the words in the input string.

\item
\textbf{Uniqueness}: Each word has only one head; that is, the dependency links
do indeed form a tree rather than some other kind of graph.

Most dependency grammars assume uniqueness, but that of Hudson \cite{HudEWG} does
not; Hudson uses multiple heads to account for transformational phenomena, where
a single word has connections to more than one position in the sentence.

\item
\textbf{Projectivity (adjacency)}:  If word A depends on word B, then all
words between A and B are also subordinate to B.  This is equivalent to
``no crossing branches'' in a constituency tree.

Some dependency grammars assume projectivity, and others do not.  In an earlier
paper \cite{Cov90} I showed how to adapt dependency parsing to a language with
totally free word order.  This of course entails abandoning projectivity.

\item
\textbf{Word-at-a-time operation}: The parser examines words one at a time,
attaching them to the tree as they are encountered, rather than waiting for
complete phrases.

This excludes dependency parsers that
are simple notational variants of constituency parsers.

\item
\textbf{Single left-right pass}: Unless forced to backtrack because of ambiguity,
the parser makes a single left-to-right pass through the input string.  This is a
vague requirement until the other requirements are spelled out more fully, but it
excludes parsers that look ahead an indefinite distance, find a head, and back up
to find its predependents (compare \cite{HeadCorner}).

\item
\textbf{Eagerness}:  The parser establishes each link as early in its left-right
pass as possible.   Abney argues convincingly that eagerness is a property
of the parsers in our heads \cite{Abney}.
\end{itemize}

\subsection{Simplifying assumptions.}

For this initial investigation I will make four more assumptions
that will definitely need to be relaxed when parsing natural language with
actual grammars.
\begin{itemize}
\item
\textbf{Instant grammar}:  I assume that the grammar can tell the parser, in
constant time, whether any given pair of words can be linked, and if so, which
is the head and which is the dependent.  (In a real grammar, some links could be
harder to work out than others.)

\item
\textbf{No ambiguity}:  I assume that there is neither local nor global ambiguity
in any parse tree; that is, every link put in place by the parser is part of the
ultimately correct parse.

This is clearly false for natural language, but by
assuming it, I can postpone consideration of how to manage ambiguity.
Psychological evidence indicates that the parsers in our heads encounter
relatively little local ambiguity, and that they backtrack when necessary
\cite{Abney}.

\item
\textbf{No inaudibilia}:  The grammar does not postulate any inaudible elements
such as null determiners, null auxiliaries, or traces.  (Bottom-up parsers
cannot respond to inaudibilia.)

\item
\textbf{Atomicity}:  I assume that words are unanalyzable elements and that there
are no operations on features or words' internal structure.

\end{itemize}

%
%

\begin{figure}
\centerline{\includegraphics[width=3.3in]{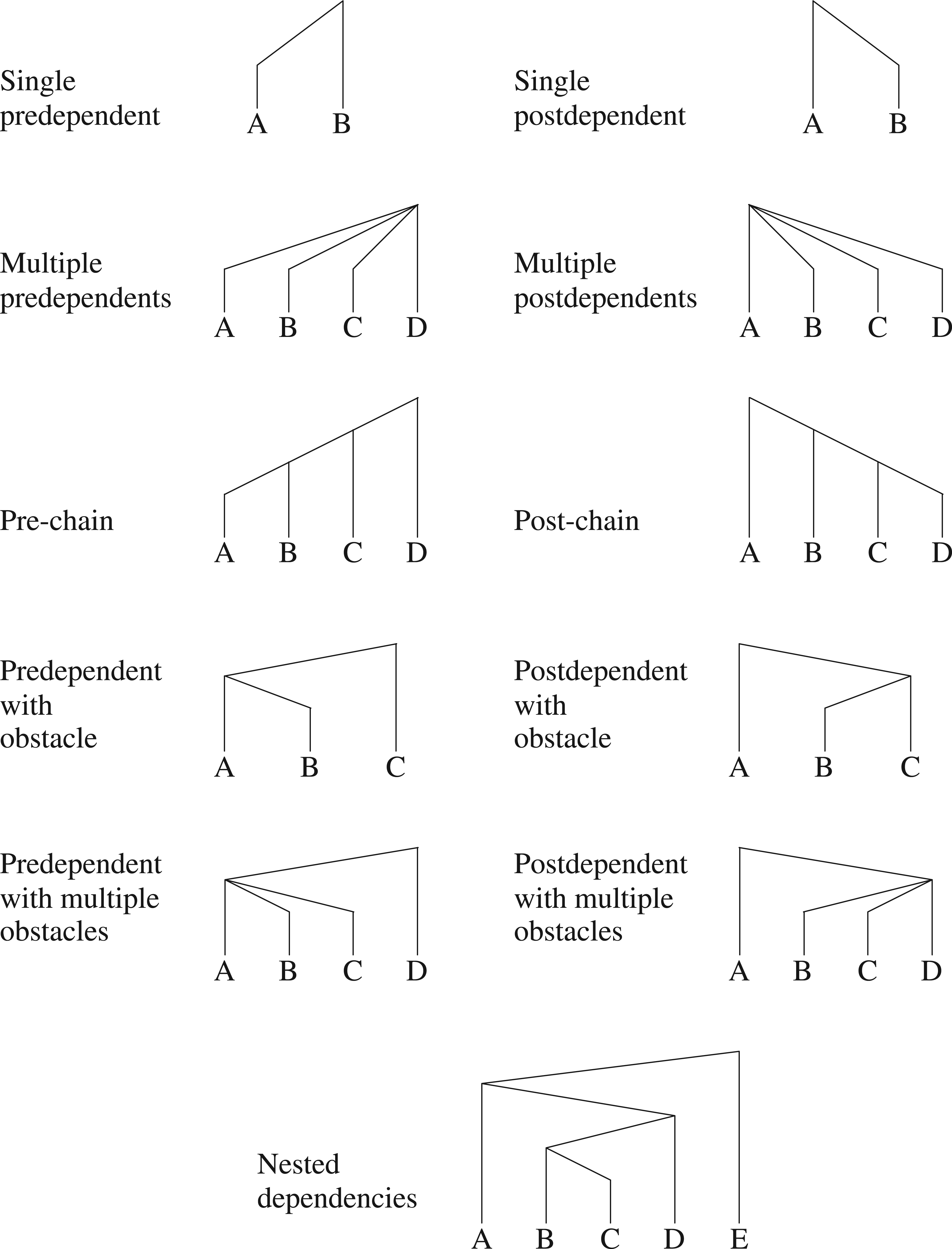}}
\caption{Some projective (non-crossing) structures that any
dependency parser should handle.}
\label{testsuit}
\end{figure}

\begin{figure}
\centerline{\includegraphics[width=3.3in]{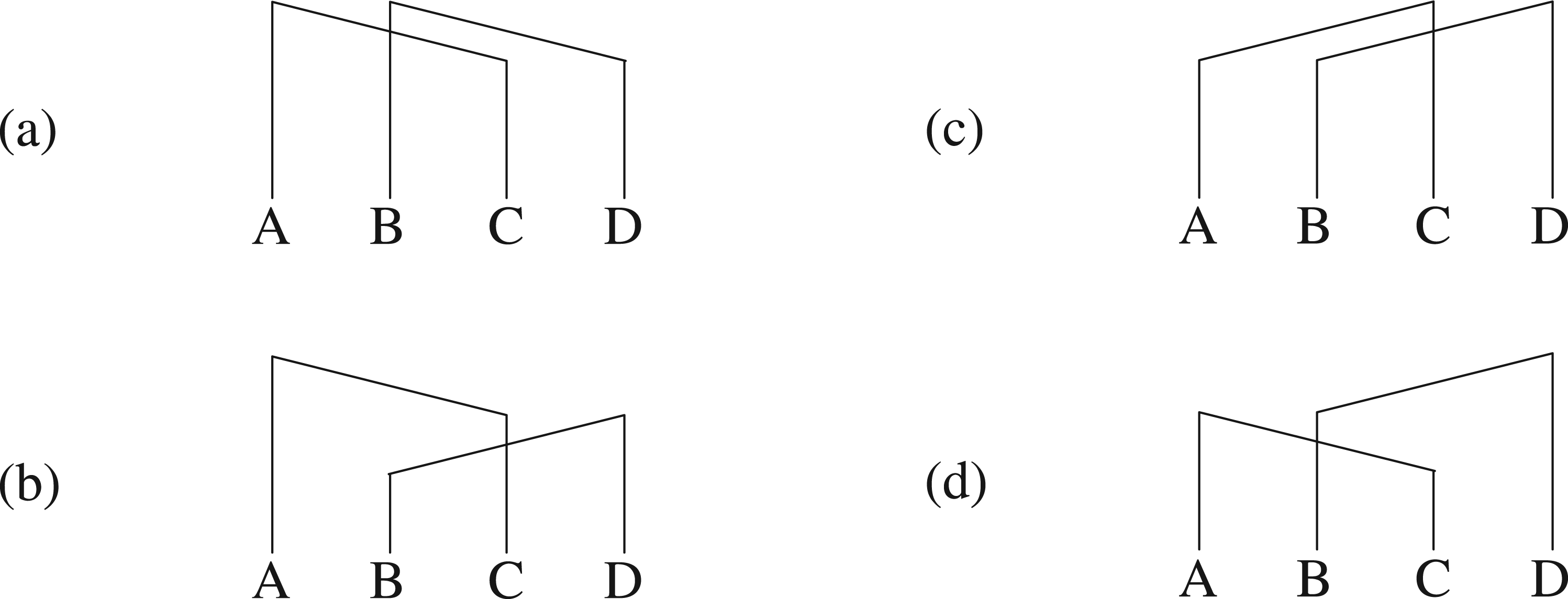}}
\caption{Some non-projective structures, allowed in some languages and not in others.}
\label{nonpro}
\end{figure}

Figures \ref{testsuit} and \ref{nonpro} show ``test suites'' of projective and
non-projective structures that parsers should handle.

\section{The obvious parsing strategy.}

Given these assumptions, one parsing strategy is obvious.  I call it a strategy and
not an algorithm because it is not yet fully specified:

\begin{strategy}[Brute-force search]
Examine each pair of words in the entire sentence, linking them as head-to-dependent
or dependent-to-head if the grammar permits.
\end{strategy}

That is, for $n$ words, try all $n(n-1)$ pairs.
Note that the number of pairs, and hence the parsing complexity, is $O(n^2)$.
If backtracking were permitted, it would be $O(n^3)$, just like constituency
parsing, because in the theoretical worst case, the whole process might have to
be done afresh after accepting each word.

Implemented as a single
left-to-right pass, the brute-force search strategy is essentially this:

\begin{strategy}[Exhaustive left-to-right search]
Accept words one by one starting at the beginning of the sentence, and try linking
each word as head or dependent of every previous word.
\end{strategy}

This still leaves the order of comparisons unspecified.  When looking for potential
links to word $n$, do we work backward, through words $n-1$, $n-2$, etc., down to $1$,
or forward, from word $1$ to word $2$ up to $n-1$?

Clearly, if the grammar enforces projectivity, or even if projective structures are
merely predominant, then the head and dependents of any given word are more likely to
be near it than far away.  Thus, they will be found earlier by working backward than
by working forward.

Whether it is better to look for heads and then dependents, or dependents and then heads,
or both concurrently, cannot yet be determined.  Thus we have two fully specified algorithms:

\begin{flushleft}\obeylines
{\bf Algorithm ESH
(Exhaustive left-to-right search, heads first)}
\textsl{Given an $n$-word sentence:}\smallskip
[1] \textbf{for} $i:=1$ \textbf{to} $n$ \textbf{do}
[2] \textbf{begin}
[3] \quad \textbf{for} $j:=i-1$ \textbf{down to} $1$ \textbf{do}
[4] \quad \textbf{begin}
[5] \quad \quad \textsl{If the grammar permits, }
\qquad\qquad\qquad \textsl{link word $j$ as head of word $i$;}
[6] \quad \quad \textsl{If the grammar permits, }
\qquad\qquad\qquad \textsl{link word $j$ as dependent of word $i$}
[7] \quad \textbf{end}
[8] \textbf{end}
\end{flushleft}

\begin{flushleft}\obeylines
{\bf Algorithm ESD
(Exhaustive left-to-right search, dependents first)}
\textsl{Same, but with steps \textup{[5]} and \textup{[6]} swapped.}
\end{flushleft}

Note that although these algorithms are expressed in terms of arrays indexed by
$i$ and $j$, they can also be implemented with linked lists or
in some other way.

\section{Refining the algorithms.}

Those na\"{\i}ve algorithms are obviously inefficient.  A better dependency parsing algorithm
should not even try links
that would violate unity, uniqueness, or (when required by the language)
projectivity.

Because the parser operates one word at a time, unity can only be checked at the end of
the whole process: did it produce a tree with a single root that comprises all of the words?
Uniqueness and projectivity, however, can and should be built into the parsing algorithm.
Here is how to handle uniqueness:

\begin{strategy}[Enforcing uniqueness]
\hfill
\begin{itemize}
\item Principle: When a word has a head, it cannot have another one.
\item Implementation:
\begin{itemize}
\item When looking for \emph{dependents} of the current word,
do not consider words that are already dependents of something else.

\item When looking for the \emph{head} of the current word,
stop after finding one head; there will not be another.
\end{itemize}
\end{itemize}
\end{strategy}
This leads immediately to:
\begin{flushleft}\obeylines
{\bf Algorithm ESHU
(Exhaustive search, heads first, with uniqueness)}
\textsl{Given an $n$-word sentence:}\smallskip
[1] \textbf{for} $i:=1$ \textbf{to} $n$ \textbf{do}
[2] \textbf{begin}
[3] \quad \textbf{for} $j:=i-1$ \textbf{down to} $1$ \textbf{do}
[4] \quad \textbf{begin}
[5] \quad \quad \textsl{If no word has been }
\qquad\qquad\qquad \textsl{linked as head of word $i$, then}
[6] \quad \quad \quad \textsl{if the grammar permits, }
\qquad\qquad\qquad \textsl{link word $j$ as head of word $i$;}
[7] \quad \quad \textsl{If word $j$ is not a dependent}
\qquad\qquad\qquad \textsl{ of some other word, then}
[8] \quad \quad \quad \textsl{if the grammar permits, }
\qquad\qquad\qquad \textsl{link word $j$ as dependent of word $i$}
[9] \quad \textbf{end}
[10] \textbf{end}
\end{flushleft}

\begin{flushleft}\obeylines
{\bf Algorithm ESDU
(Exhaustive search, dependents first, with uniqueness)}
\textsl{Same, but with \textup{[5--6]} and \textup{[7--8]} swapped.}
\end{flushleft}

Here the advantages of a list-based representation begin to become apparent.  Rather than work
through an array and perform tests to determine which elements to skip, it is simpler
to work through lists from which the ineligible elements have already been removed.

Here is an algorithm that works with two lists, \emph{Wordlist} and \emph{Headlist},
containing,
respectively, all the words encountered so far and all the words that lack heads.
Both lists are built by adding elements at the beginning, so they contain words in the
opposite of the order in which they were encountered.  As a result, searching each
list from the beginning retrieves the most recent words first.

\begin{flushleft}\obeylines
{\bf Algorithm LSU
(List-based search with uniqueness)}
\textsl{Given a list of words to be parsed,
and two working lists \textit{Headlist} and \textit{Wordlist}:}\smallskip
{(Initialize)}
$\textit{Headlist} := [];$ \qquad {(Words that do not yet have heads)}
$\textit{Wordlist} := [];$ \qquad {(All words encountered so far)}\smallskip
\textbf{repeat}\smallskip
\qquad {(Accept a word and add it to \textit{Wordlist})}
\qquad $W := \textsl{the next word to be parsed};$
\qquad $\textit{Wordlist} := W + \textit{Wordlist}$;\smallskip
\qquad {(Dependents of $W$ can only be in \textit{Headlist})}
\qquad \textbf{for} $D :=$ \textsl{each element of \textit{Headlist}, }
\qquad\qquad\qquad \textsl{starting with the first}
\qquad\quad \textbf{begin}
\qquad\quad\quad \textbf{if} \textsl{$D$ can depend on $W$} \textbf{then}
\qquad\quad\quad\quad \textbf{begin}
\qquad\quad\quad\quad\quad \textsl{link $D$ as dependent of $W$};
\qquad\quad\quad\quad\quad \textsl{delete $D$ from \textit{Headlist}}
\qquad\quad\quad\quad \textbf{end}
\qquad\quad \textbf{end};\smallskip
\qquad {(Look for the head of $W$; there can only be one)}
\qquad \textbf{for} $H :=$ \textsl{each element of \textit{Wordlist}, }
\qquad\qquad\qquad  \textsl{starting with the first}
\qquad\quad \textbf{if} \textsl{$W$ can depend on $H$} \textbf{then}
\qquad\quad\quad \textbf{begin}
\qquad\quad\quad\quad \textsl{link $W$ as dependent of $H$};
\qquad\quad\quad\quad \textsl{terminate this \textbf{for} loop}
\qquad\quad\quad \textbf{end};
\qquad \textbf{if} \textsl{no head for $W$ was found} \textbf{then}
\qquad\quad $\textit{Headlist} := W + \textit{Headlist};$\smallskip
\textbf{until} \textsl{all words have been parsed.}
\end{flushleft}

This time dependents are sought before seeking heads.  The reason is that $W$, the current
word, is itself added to \emph{Headlist} if it has no head, and a step is saved by not
doing this until the search of \emph{Headlist} for potential dependents of $W$ is complete.
This is essentially the algorithm of my earlier paper \cite{Cov90}.

\section{Projectivity.}

\subsection{Definition.}

Projectivity is informally defined as ``no crossing branches.'' More formally:
\begin{itemize}
\item
A tree is projective if and only if every word in it comprises a continuous substring.
\item
A word comprises a continuous substring if and only if,
given any two words that it comprises,
it also comprises all the words between them.
\end{itemize}
The second clause of this is simply the definition of ``continuous'' -- a continuous
substring is one such that everything between any of its elements is also part of it.


\subsection{Building projectivity into the parser.}

Now how does all of this apply to parsing?  To build projectivity into a
bottom-up dependency parser,
we need to constrain it as follows:
\begin{itemize}
\item[(a)] Do not skip a potential predependent of $W$.  That is, either
attach every consecutive preceding word that is still independent, or stop searching.

\item[(b)] When searching for the head of $W$, consider only the 
\textsf{\textbf{most recent word not already subordinate to $W$,}}
its head, that word's head, and so on to the root of the tree.\footnote{\textsf{\textbf{In the
2001 version of the paper, I said, erroneously, ``consider only the previous word,''
which however might be subordinate to $W$, in which case the algorithm 
would simply climb up to $W$ and proceed no further.
Such an algorithm
could not
parse a postdependent with
multiple obstacles as shown in Fig.\ 5.}}}
\end{itemize}

Constraint (b) is easy to understand.
It says that if the head of $W$ (call it $H$) precedes $W$, it must 
\textsf{\textbf{be, or 
comprise, the most recent word preceding $W$ that is not already
subordinate to $W$}}; thus, it is reachable by climbing the
tree from that word.  This follows from the definition
of projectivity: \textsf{\textbf{everything from $H$ to $W$ must
be a continuous substring}}.

Constraint (a) says that the predependents of $W$ are a continuous string of
the words that are still independent at the time $W$ is encountered.

Consider the words that, at any stage, still do not have heads, i.e.,
the contents of \emph{Headlist} in the list-based parsing algorithm.
Each such word is the head of a constituent, i.e., a continuous
substring.  That is, each still-independent word
stands for the string of words that it comprises.
The goal of the parser is to assemble zero or more of these strings into a
continuous string that ends with $W$.  Clearly, if any element is skipped,
the resulting string cannot be continuous.  \textsc{q.e.d.}

Here is the list-based parsing algorithm with projectivity added.
This algorithm was mentioned briefly in \cite{Cov90}.

\begin{flushleft}\obeylines
{\bf Algorithm LSUP
(List-based search with uniqueness and projectivity)}
\textsl{Given a list of words to be parsed,
and two working lists \textit{Headlist} and \textit{Wordlist}:}\smallskip
{(Initialize)}
$\textit{Headlist} := [];$ \qquad {(Words that do not yet have heads)}
$\textit{Wordlist} := [];$ \qquad {(All words encountered so far)}\smallskip
\textbf{repeat}\smallskip
\qquad {(Accept a word and add it to \textit{Wordlist})}
\qquad $W := \textsl{the next word to be parsed};$
\qquad $\textit{Wordlist} := W + \textit{Wordlist}$;\smallskip
\qquad (Look for dependents of $W$; they can only be
\qquad \quad consecutive elements of \emph{Headlist}
\qquad \quad starting with the most recently added)
\qquad \textbf{for} $D :=$ \textsl{each element of \textit{Headlist},
\qquad\qquad\qquad starting with the first}
\qquad\quad \textbf{begin}
\qquad\quad\quad \textbf{if} \textsl{$D$ can depend on $W$} \textbf{then}
\qquad\quad\quad\quad \textbf{begin}
\qquad\quad\quad\quad\quad \textsl{link $D$ as dependent of $W$};
\qquad\quad\quad\quad\quad \textsl{delete $D$ from \textit{Headlist}}
\qquad\quad\quad\quad \textbf{end}
\qquad\quad\quad \textbf{else}
\qquad\quad\quad\quad \textsl{terminate this \textbf{for} loop}
\qquad\quad \textbf{end};\smallskip
\qquad {(Look for the head of $W$; it must
\qquad\qquad\qquad comprise the word preceding $W$)}
\qquad $H :=$ \textsl{the word \textup{\textsf{\textbf{\small most recently preceding $W$}}} in
\qquad\qquad\qquad the input string \textup{\textsf{\textbf{\small that is not already}}}
\qquad\qquad\qquad \textup{\textsf{\textbf{\small subordinate to}}} $W$};
\qquad\textbf{loop}
\qquad\quad \textbf{if} \textsl{$W$ can depend on $H$} \textbf{then}
\qquad\quad\quad \textbf{begin}
\qquad\quad\quad\quad \textsl{link $W$ as dependent of $H$};
\qquad\quad\quad\quad \textsl{terminate the loop}
\qquad\quad\quad \textbf{end};
\qquad\quad\textbf{if} $H$ \textsl{is independent} \textbf{then} \textsl{terminate the loop};
\qquad\quad $H := \textsl{the head of~} H$
\qquad \textbf{end loop};
\qquad \textbf{if} \textsl{no head for $W$ was found} \textbf{then}
\qquad\quad $\textit{Headlist} := W + \textit{Headlist};$\smallskip
\textbf{until} \textsl{all words have been parsed.}
\end{flushleft}

\section{Complexity.}

We saw already that the complexity of the initial, brute-force search algorithm,
with a completely deterministic grammar, is $O(n^2)$ because the search involves
$n(n-1)$ pairs of words, and $n(n-1)$ approaches $n^2$ as $n$ becomes large.

So far I have not introduced any mechanism for handling local ambiguity.
The obvious way to do so is to \textbf{backtrack} -- that is, return to the most
recent untried alternative whenever an alernative is needed.  If the parser
is implemented in Prolog, backtracking is provided automatically.

The complexity of brute-force-search parsing with backtracking is $O(n^3)$ because,
after each of the $n$ words is accepted, the whole $O(n)$ process may have to be done
over from the beginning.  $O(n^3)$ is also the complexity of recursive-descent
constituency parsing.

\begin{figure}
\centerline{\includegraphics[width=2.5in]{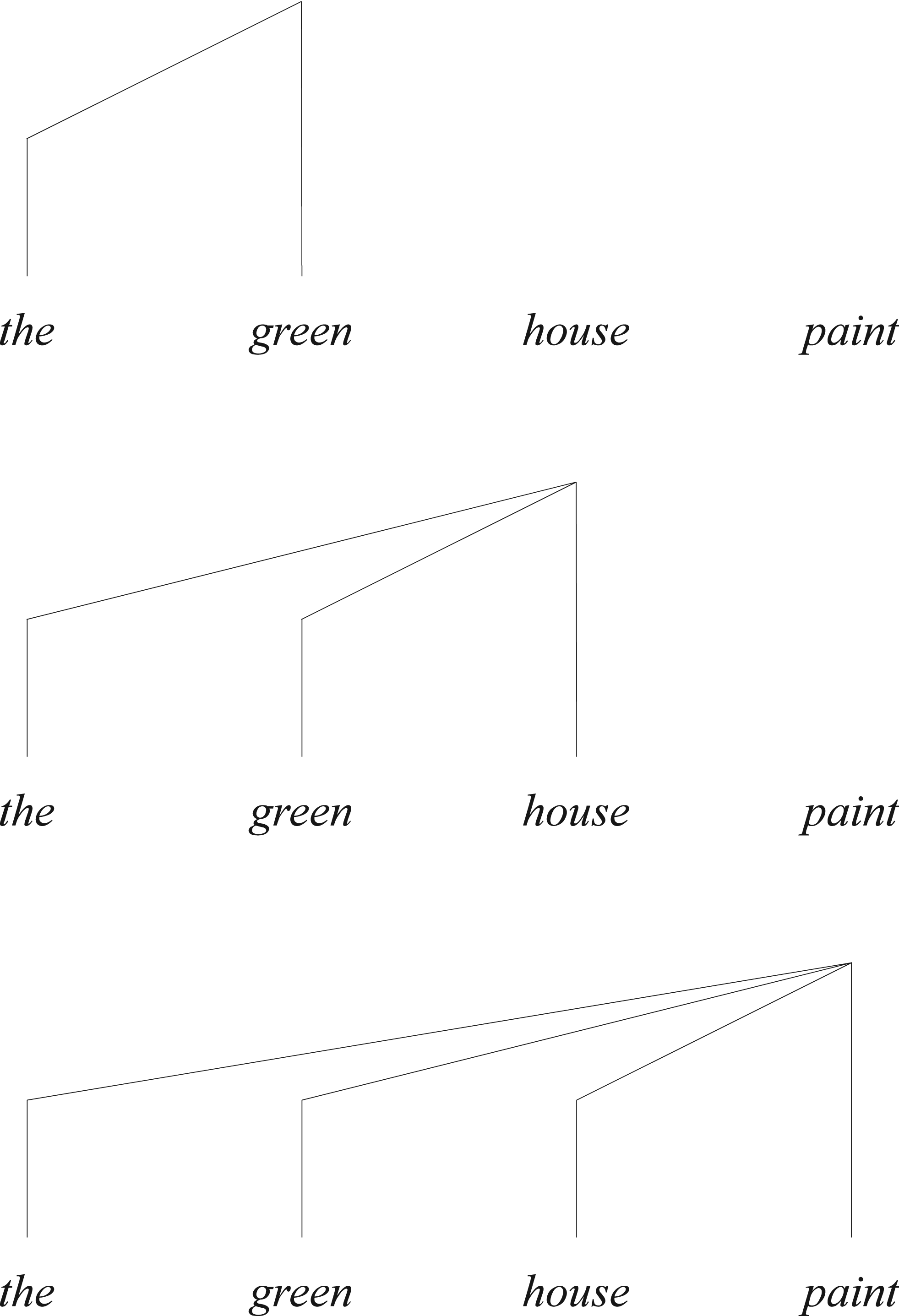}}
\caption{An instance of worst-case parsing complexity: after accepting each word,
the parser has to rework the entire structure.}
\label{wrstcase}
\end{figure}

These complexity results are not affected by constraints to enforce unity and
projectivity, since there are cases in which these constraints do not shorten
the parsing process.  Consider for example the local ambiguity in the phrase
\emph{the green house paint}.  Not only is \emph{the green} a valid phrase
(as in ``you forgot the green,'' said to a painter), but so are \emph{the green
house} and \emph{the green house paint}.  Thus, the parser must backtrack on
accepting each successive word (Fig. \ref{wrstcase}).

At this point I am still assuming atomicity.  Barton, Berwick and Ristad \cite{BBer}
prove that when lexical ambiguity and agreement features are present --- that is, when
words can be ambiguous and can be labeled with attributes --- natural
language parsing is NP-complete.

Bear in mind that these are worst-case results.  An important principle of linguistics
seems to be that \textbf{the worst case does not occur,} i.e., people do not actually
utter sentences that put any reasonable parsing algorithm into a worst-case situation.
Human language does not use unconstrained phrase-structure or dependency grammar;
it is constrained in ways that are still being discovered.

\end{document}